\documentclass[times, review, 10pt]{elsarticle}

\usepackage[top=4.3cm,right=4.8cm,bottom=4.3cm,left=4.8cm]{geometry}
\usepackage{amsmath}
\usepackage{amssymb}
\usepackage{booktabs}
\usepackage{multirow}
\usepackage{algorithm}
\usepackage{algorithmic}
\usepackage{xcolor}
\usepackage{graphicx}
\usepackage{url}
\usepackage{setspace}
\usepackage{caption}
\usepackage[hidelinks]{hyperref}

\graphicspath{{figs/}}
\renewcommand{\footnotesize}{\fontsize{8pt}{9.5pt}\selectfont}
\newcommand{\changed}[1]{\textcolor{black}{#1}}
\newcommand{\figref}[1]{\NoHyper\textcolor{black}{Fig.\ref{#1}}\endNoHyper}

\newcommand{\TableFont}{\normalsize}

\journal{Pattern Recognition}

\begin{document}

\begin{frontmatter}

\title{Learning Molecular Representations from Cellular Phenotypes with Structure Preservation}

\author[xtu]{Xuan Lin}
\author[xtu]{Jingyu Sheng}
\author[hnu]{Tengfei Ma\corref{cor1}}
\ead{tfma@hnu.edu.cn}
\author[bupt]{Li Sun\corref{cor1}}
\ead{lsun@bupt.edu.cn}
\author[seu]{Dapeng Xiong}

\cortext[cor1]{\changed{Corresponding authors.}}

\address[xtu]{\changed{School of Computer Science, Xiangtan University, Xiangtan, Hunan, China}}
\address[hnu]{\changed{College of Computer Science and Electronic Engineering, Hunan University, Changsha, Hunan, China}}
\address[bupt]{\changed{School of Computer Science, Beijing University of Posts and Telecommunications, Beijing, China}}
\address[seu]{\changed{State Key Laboratory of Digital Medical Engineering, School of Biological Science and Medical Engineering, Southeast University, Nanjing, Jiangsu, China}}

\begin{abstract}
Phenotypic drug discovery enables the discovery of functional relationships between molecular structures and cellular responses. However, existing multimodal representation learning methods often optimize cross-modal alignment without considering the intrinsic organization of chemical space, resulting in distorted molecular representations and loss of structural information.
We propose \textbf{PhenMol}, a structure-preserving framework for phenotype-aware molecular representation learning. PhenMol disentangles molecular and cellular representations into shared and private components, enabling phenotype-guided alignment while preserving chemical structures through a dedicated molecular branch. This design integrates cellular phenotype information without disrupting molecular neighborhood organization.
Experiments on approximately $3.04 \times 10^{4}$ molecule--cell morphology pairs demonstrate that PhenMol improves molecular property prediction across 270 bioactivity tasks, molecule--phenotype retrieval, and clinical trial outcome prediction. Moreover, ECFP4-based structural analysis shows that PhenMol better preserves molecular neighborhoods and reduces embedding distortion compared with existing multimodal alignment methods.
These results highlight the importance of structure-aware constraints in multimodal molecular representation learning and provide an effective approach for integrating cellular phenotypes with chemical knowledge for drug discovery.

\end{abstract}

\begin{keyword}
\changed{Phenotypic drug discovery \sep Molecular representation learning \sep Cell Painting \sep Multimodal learning \sep Bioactivity prediction}
\end{keyword}

\end{frontmatter}

\section{Introduction}

Molecular representation learning underpins computational drug discovery~\cite{computational_drug_discovery:1}, supporting tasks such as virtual screening~\cite{virtual_screening:1}, property prediction~\cite{pattern_recognition:lemon}, target identification~\cite{target_identification:1}, and clinical outcome assessment~\cite{clinical_outcome_assessment:1}. Existing approaches, including molecular fingerprints~\cite{ecfp:0}, graph neural networks~\cite{gnn:0}, and SMILES-based models~\cite{smiles:0}, primarily learn from intrinsic chemical structure and have been strengthened by self-supervised pretraining, \textcolor{black}{including contextual property and motif prediction in GROVER \cite{Self_Supervised:grover}, graph augmentation in GraphCL \cite{Self_Supervised:graphcl} and JOAO \cite{Self_Supervised:joao}, molecular graph contrastive learning in MolCLR \cite{Self_Supervised:molclr}, and knowledge-guided masked graph modeling in KPGT \cite{Self_Supervised:kpgt}.} Although these representations provide strong chemical priors, they do not directly capture how molecules perturb living cells, motivating the integration of cellular phenotypes into molecular representation learning.

Phenotypic drug discovery offers a complementary route by directly profiling cellular responses to small molecules, enabling the discovery of therapeutically relevant phenotypes and mechanisms that may be difficult to infer from chemical structure alone~\cite{9}. High-content imaging platforms such as Cell Painting~\cite{cellpainting:1,cellpainting:2} capture morphological changes~\cite{morphological:1,morphological:2} across multiple cellular compartments and provide rich functional readouts of compound action. These advances have motivated phenotype-aware molecular representation learning, which integrates cellular phenotypes with molecular structure to produce biologically informed representations for drug discovery~\cite{baseline:moshkov}. The key challenge is to exploit this biological context without disrupting the chemical organization required for molecular generalization.

\begin{figure*}[t]
    \centering
    \includegraphics[width=\textwidth]{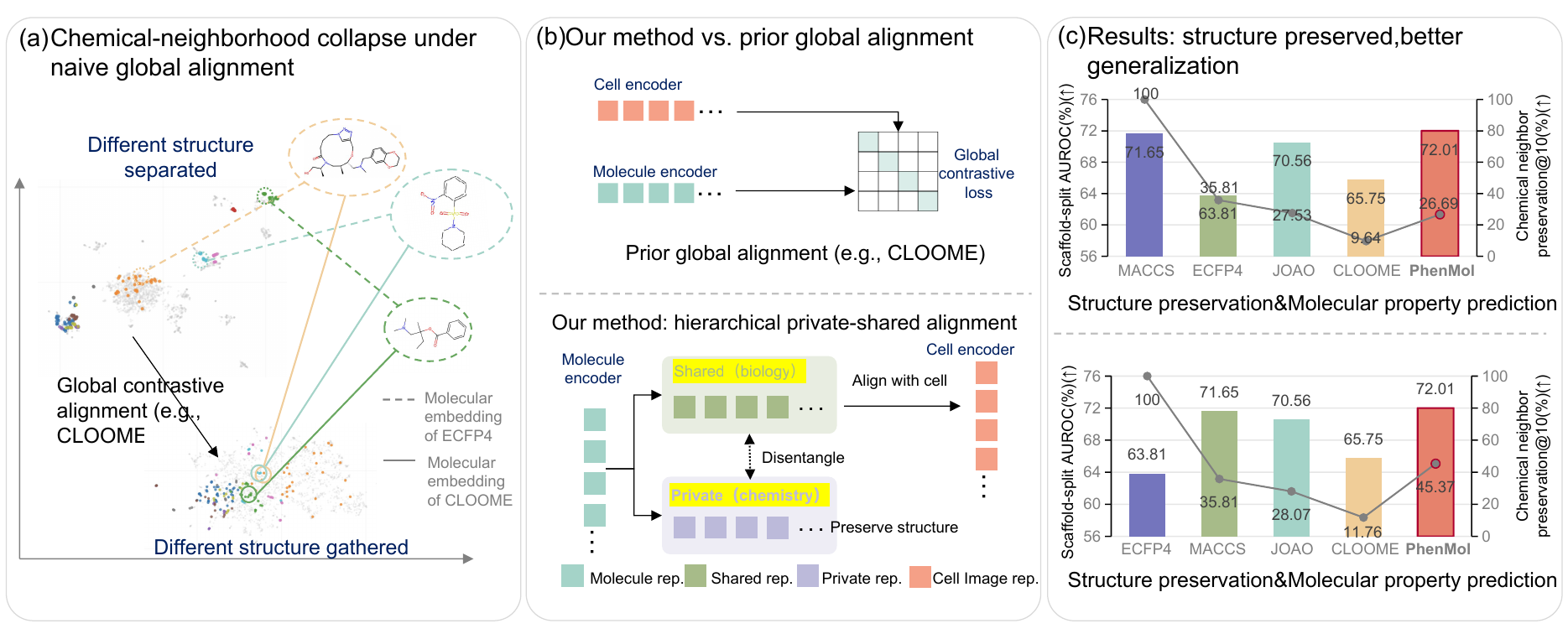}
    \caption{Motivation for structure-preserving phenotype-aware molecular representation learning.
(a) Naive global contrastive alignment can cause \emph{molecular structure collapse}, disrupting chemically meaningful local organization under noisy phenotypic supervision.
(b) PhenMol separates shared cross-modal signals from modality-private chemical and image factors to preserve molecular structure during alignment.
(c) PhenMol quantitatively retains more ECFP4-defined chemical neighbors and improves scaffold-split bioactivity prediction, showing that structural preservation is critical for structure-sensitive generalization.}
    \label{fig:intro_motivation}
\end{figure*}

However, incorporating cellular phenotypes into molecular representation learning creates a fundamental tension between biological alignment and chemical fidelity. Existing methods typically align molecular and cellular embeddings through global contrastive objectives \cite{14,baseline:clip}. \textcolor{black}{Recent work on molecular graph contrastive learning~\cite{pattern_recognition:lemon} and biomedical image--text fusion~\cite{pattern_recognition:medical_fusion} further highlights the broader need to learn informative representations while controlling cross-view interference.} Yet cellular images are high-dimensional and continuous measurements affected by batch effects, staining variation, cell density, and imaging artifacts \cite{15,16}, whereas molecular representations encode discrete structural priors defined by scaffolds, functional groups, and topology \cite{ecfp:0,activity_cliffs:1}. Forcing these heterogeneous modalities into a shared space can cause molecular embeddings to over-assimilate phenotype-specific variation, disrupting the organization of the chemical representation space. Chemically similar molecules may drift apart, while structurally distinct molecules may become artificially close. We refer to this failure mode as \emph{molecular structure collapse} (\figref{fig:intro_motivation}a), in which a biological context is incorporated at the expense of chemical fidelity.

Existing phenotype-aware methods improve molecule--cell correspondence but do not explicitly prevent this collapse. CLOOME~\cite{baseline:cloome} relies primarily on global molecule--image contrastive learning, while MIGA~\cite{baseline:MIGA} and MINER~\cite{baseline:MINER} introduce auxiliary matching, robustness, or noise-mitigation objectives. \textcolor{black}{Related multimodal representation studies also show that cross-view fusion and contrastive alignment can improve biomedical pattern recognition when modality-specific interference is controlled~\cite{pattern_recognition:medical_fusion,pattern_recognition:focuscontrast}.} Although these strategies strengthen cross-modal learning, they mainly optimize alignment quality and do not explicitly separate phenotype-related signals from intrinsic molecular structure (\figref{fig:intro_motivation}b). Consequently, preserving chemically meaningful molecular organization during phenotype integration remains an open challenge.

To address this challenge, we introduce \textbf{PhenMol}, a structure-preserving framework for phenotype-aware molecular representation learning. PhenMol decomposes molecular and cellular representations into modality-shared and modality-private components, discourages batch-specific variation from dominating the shared cellular space, and preserves scaffold-level chemical information in a dedicated molecular private branch. Combined with contrastive alignment and latent-space pair matching, this design integrates cellular phenotypic signals while maintaining chemically meaningful molecular organization.
We evaluate PhenMol on approximately $3.04 \times 10^{4}$ paired molecule--cell morphology samples across molecular property prediction, bidirectional molecule--image retrieval, clinical trial outcome prediction, ablation studies, and structural preservation analysis. As shown in \figref{fig:intro_motivation}c, PhenMol preserves substantially more ECFP4-defined molecular neighbors than prior global alignment methods while achieving the best scaffold-split bioactivity prediction performance. This result shows that maintaining chemical organization during phenotype integration directly benefits structure-sensitive generalization, rather than serving only as an auxiliary representation constraint.
The main contributions of this work are:
\begin{itemize}
    \item We identify \textbf{molecular representation collapse under phenotype alignment} as a critical challenge in phenotype-aware molecular representation learning, where global multimodal alignment can compromise chemical neighborhood preservation.
    \item We introduce \textbf{PhenMol}, a structure-preserving multimodal framework that disentangles shared biological signals from private chemical information to integrate cellular phenotypes without sacrificing molecular identity.
    \item We demonstrate through molecular prediction, retrieval, clinical transfer, and structural analyses that \textbf{PhenMol} learns phenotype-aware molecular representations that are both biologically informative and chemically faithful.
\end{itemize}

\section{Related Work}

\subsection{Phenotype-Aware Molecular Representation Learning}

Molecular representations based on fingerprints, graph neural networks, and molecular Transformers provide strong chemical priors for drug discovery \textcolor{black}{and have been further improved through self-supervised objectives, including motif prediction in GROVER~\cite{Self_Supervised:grover}, graph augmentation in GraphCL~\cite{Self_Supervised:graphcl} and JOAO~\cite{Self_Supervised:joao}, molecular graph contrastive learning in MolCLR~\cite{Self_Supervised:molclr} and LEMON~\cite{pattern_recognition:lemon}, knowledge-guided masked graph modeling in KPGT~\cite{Self_Supervised:kpgt}, structure-enhanced protein--ligand representation learning~\cite{pattern_recognition:ksm}, and multi-view learning with structured or unstructured knowledge~\cite{molecule_multiview:mvmol}.} However, these approaches mainly characterize intrinsic molecular structure and do not directly capture how compounds perturb living cells.
Phenotypic drug discovery complements structure-based modeling by measuring compound-induced biological responses. High-content imaging and Cell Painting~\cite{cellpainting:1,cellpainting:2} capture morphology changes~\cite{morphological:1,morphological:2} across cellular compartments, while transcriptomic profiles such as L1000 provide pathway-level response signatures~\cite{baseline:moshkov}. These readouts can reveal drug effects that are difficult to infer from chemical structure alone, but they also contain batch effects, staining variation, cell-density differences, and other assay-specific noise. The resulting challenge is therefore not merely to incorporate cellular phenotypes into molecular representations, but to exploit their biological information without distorting the chemical organization required for molecular similarity, activity prediction, and scaffold-level generalization. This motivates PhenMol, which explicitly balances phenotype integration with molecular structure preservation.

\subsection{Cell--Molecule Alignment with Structural Preservation}

Existing phenotype-aware methods typically align molecular and cellular representations in a shared latent space. CLOOME~\cite{baseline:cloome} adopts CLIP-style~\cite{baseline:clip} global image--molecule contrastive learning. MIGA~\cite{baseline:MIGA} introduces auxiliary matching or reconstruction objectives, and MINER~\cite{baseline:MINER} improves robustness to noisy phenotypic supervision through calibrated alignment strategies. \textcolor{black}{Related biomedical multimodal learning studies further show that image--text fusion and attention-guided contrastive views can improve representation quality under heterogeneous inputs~\cite{pattern_recognition:medical_fusion,pattern_recognition:focuscontrast}.} These methods demonstrate the value of cellular morphology for molecular representation learning.
Nevertheless, most prior work primarily optimizes cross-modal similarity or robustness, without explicitly preserving the structural geometry of the molecular embedding space. Because cellular images contain batch effects, staining variation, and other phenotype-specific noise, unrestricted alignment may cause chemically similar molecules to drift apart and structurally distinct molecules to become artificially close. PhenMol addresses this limitation by separating modality-shared biological information from modality-private features and preserving scaffold-level chemical priors during alignment. It therefore learns molecular representations that remain chemically faithful while incorporating informative cellular responses.

\section{Method}

\begin{figure*}[t]
    \centering
    \includegraphics[width=\textwidth]{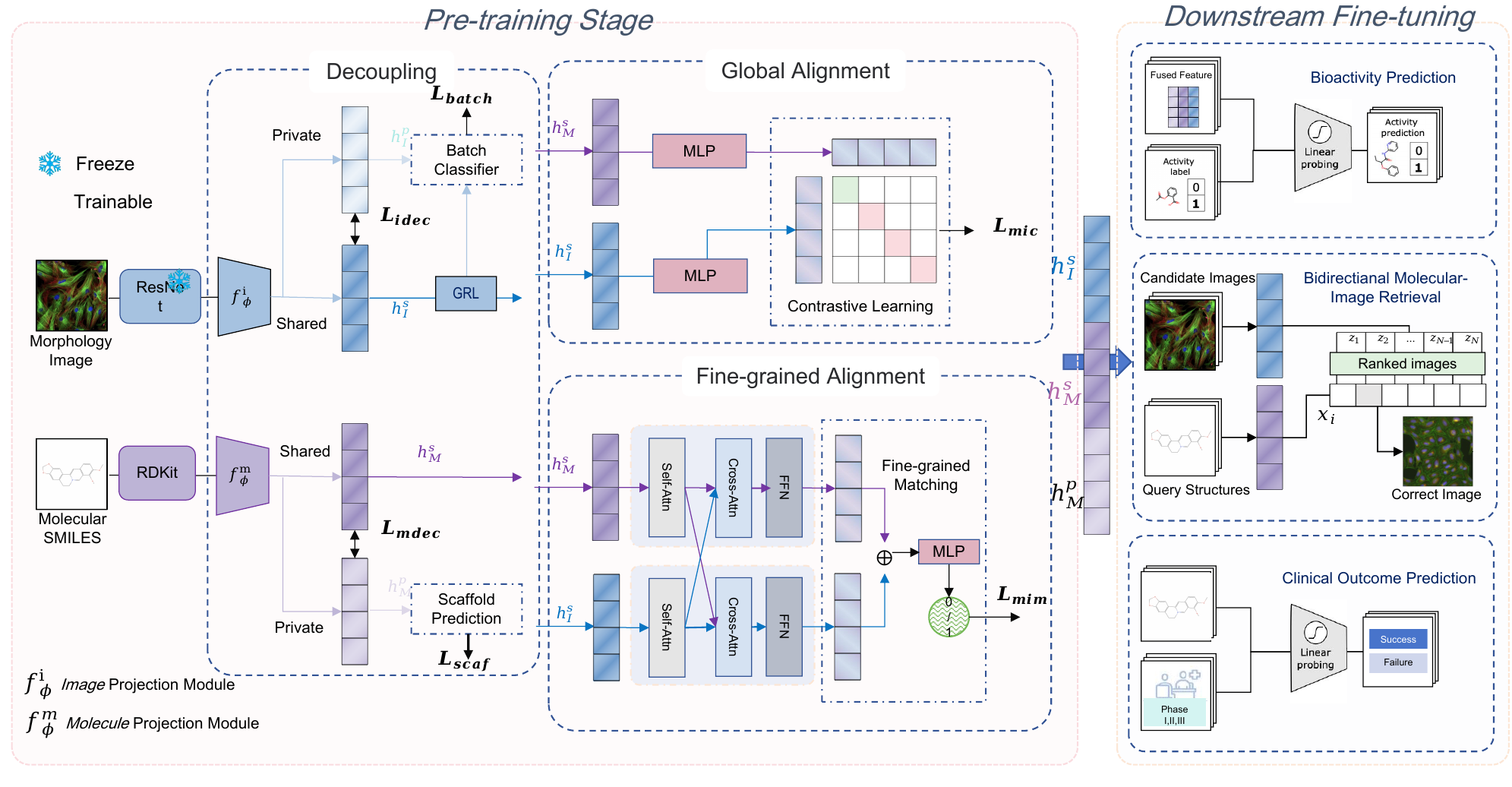}
    \caption{Overview of PhenMol.
During pre-training, PhenMol projects molecular fingerprints and cellular phenotype features into a common latent space and decomposes each modality into shared and private representations. The shared branches support global molecule-phenotype contrastive alignment and fine-grained pair matching. In contrast, the private branches retain modality-specific information, including scaffold-level chemical structure and batch-related cellular variation. The resulting representations are transferred to downstream bioactivity prediction, bidirectional molecule-image retrieval, and clinical trial outcome prediction.}
    \label{fig:modal_thir}
\end{figure*}

We propose \textbf{PhenMol}, a structure-preserving multimodal framework for molecular representation learning in phenotypic drug discovery. Given paired molecular inputs and cellular phenotype features, PhenMol integrates drug-induced biological responses into molecular representations while retaining chemically meaningful structural priors. As shown in \figref{fig:modal_thir}, PhenMol consists of three components: shared--private feature decoupling, global molecule--phenotype alignment, and fine-grained latent pair matching.

\subsection{Problem Formulation}

\textcolor{black}{Given paired molecular structures and cellular phenotypes, PhenMol aims to learn a molecular representation that is both phenotype-aware and chemically faithful. The paired pre-training dataset is defined as}
\begin{equation}
\mathcal{D}=\{(\boldsymbol{v}_i,\boldsymbol{m}_i)\}_{i=1}^{N},
\end{equation}
\textcolor{black}{where $\boldsymbol{v}_i \in \mathbb{R}^{d_v}$ denotes the pre-extracted cellular phenotype feature and $\boldsymbol{m}_i$ denotes the corresponding molecule. Each molecule $\boldsymbol{m}_i$ is converted into a 2048-dimensional ECFP4 fingerprint $\boldsymbol{x}_i^{m} \in \{0,1\}^{2048}$ using RDKit. PhenMol first learns modality-specific encoders}
\begin{equation}
\boldsymbol{z}^{v}_i=f_v(\boldsymbol{v}_i), \qquad
\boldsymbol{z}^{m}_i=f_m(\boldsymbol{x}_i^{m}),
\end{equation}
\textcolor{black}{where $\boldsymbol{z}^{v}_i,\boldsymbol{z}^{m}_i \in \mathbb{R}^{d}$ are cellular and molecular latent embeddings. The first requirement is paired cross-modal alignment: the phenotype embedding and molecular embedding from the same perturbation should be closer than mismatched pairs. This can be written as}
\begin{equation}
\operatorname{sim}(\boldsymbol{z}^{v}_i,\boldsymbol{z}^{m}_i)
>
\operatorname{sim}(\boldsymbol{z}^{v}_i,\boldsymbol{z}^{m}_j),
\qquad i\neq j .
\end{equation}
\textcolor{black}{The second requirement is molecular structure preservation: if two molecules are structurally similar under ECFP4-defined similarity, their learned molecular embeddings should also remain close. We express this relationship as}
\begin{equation}
\operatorname{sim}_{\mathrm{ECFP4}}(\boldsymbol{x}^{m}_i,\boldsymbol{x}^{m}_j)
\uparrow
\Rightarrow
\operatorname{sim}(\boldsymbol{z}^{m}_i,\boldsymbol{z}^{m}_j)
\uparrow .
\end{equation}
\textcolor{black}{The overall learning problem therefore balances phenotype alignment and chemical preservation:}
\begin{equation}
\min_{\Theta}\;
\mathcal{J}
=
\mathcal{L}_{\mathrm{align}}
+
\lambda_{\mathrm{chem}}
\mathcal{L}_{\mathrm{chem}},
\end{equation}
\textcolor{black}{where $\Theta$ denotes all learnable parameters, $\mathcal{L}_{\mathrm{align}}$ optimizes paired molecule--phenotype correspondence, $\mathcal{L}_{\mathrm{chem}}$ constrains the molecular embedding space to retain chemical organization, and $\lambda_{\mathrm{chem}}$ controls the trade-off between the two goals. After pre-training, the learned representations are used for three downstream settings: bioactivity prediction, molecule--phenotype retrieval, and clinical result prediction.}

\subsection{Framework Overview}

As illustrated in \figref{fig:modal_thir}, PhenMol first projects molecular fingerprints and cellular phenotype features into a common latent space, and then decomposes each modality into shared and private representations. The shared branches capture cross-modal biological signals, while the private molecular branch preserves chemical structure and the private cellular branch retains modality-specific variation, including batch-related effects.
PhenMol performs alignment at two complementary levels. Global contrastive learning aligns paired molecular and cellular representations in the shared space, whereas latent pair matching further models their instance-level compatibility through bidirectional interaction. The resulting representations are used for downstream molecular prediction and molecule-phenotype retrieval.

\subsection{Pre-training Stage}

\textcolor{black}{The pre-training stage is designed to integrate cellular phenotype information without allowing global alignment to overwrite chemically meaningful molecular organization. The process starts from the projected molecular and cellular embeddings, $\boldsymbol{z}^{m}$ and $\boldsymbol{z}^{v}$, and first decouples each modality into shared and private representations. The shared branches provide the interface for phenotype--molecule correspondence, whereas the private branches retain modality-specific factors, including molecular scaffold information and cellular batch-related variation. Global alignment is then applied to the shared molecular and cellular representations so that paired perturbations are close at the representation level. Finally, fine-grained matching further examines each candidate molecule--phenotype pair after latent interaction, providing instance-level compatibility evidence beyond global similarity. In this way, decoupling defines what information can be aligned, global alignment organizes the shared cross-modal space, and fine-grained matching refines pair-level correspondence while the private branches constrain structure and nuisance variation.}

\noindent \textbf{\textcolor{black}{Decoupling.}}
\textcolor{black}{Given projected cellular and molecular features, $\boldsymbol{z}^{v}$ and $\boldsymbol{z}^{m}$, PhenMol decomposes each modality into shared and private representations:}
\begin{equation}
\boldsymbol{h}_v^{s},\boldsymbol{h}_v^{p}=g_v(\boldsymbol{z}^{v}), 
\qquad
\boldsymbol{h}_m^{s},\boldsymbol{h}_m^{p}=g_m(\boldsymbol{z}^{m}).
\end{equation}
\textcolor{black}{The shared representations, $\boldsymbol{h}_v^{s}$ and $\boldsymbol{h}_m^{s}$, capture biological signals that are comparable across modalities, whereas the private representations, $\boldsymbol{h}_v^{p}$ and $\boldsymbol{h}_m^{p}$, retain modality-specific information. This separation is necessary because cell morphology contains useful perturbation responses but also batch effects and phenotype-specific nuisance variation. The molecular private branch reserves capacity for scaffold-level chemical information, while the cellular private branch absorbs batch-related variation that should not define molecular similarity. To reduce redundancy between the two spaces, we impose an orthogonality constraint:}
\begin{equation}
\mathcal{L}_{\mathrm{ortho}}
=
\ell(\boldsymbol{h}_v^{s},\boldsymbol{h}_v^{p})
+
\ell(\boldsymbol{h}_m^{s},\boldsymbol{h}_m^{p}),
\end{equation}

\textcolor{black}{where $\ell(\cdot,\cdot)$ is the average squared inner product between normalized feature vectors. For cellular phenotypes, a private batch classifier encourages batch-related variation to remain in $\boldsymbol{h}_v^{p}$, while a gradient-reversal classifier discourages such information from dominating $\boldsymbol{h}_v^{s}$~\cite{domain_adversarial:grl}. For molecules, a scaffold prediction objective is applied to $\boldsymbol{h}_m^{p}$ to preserve scaffold-level chemical information.}

\noindent \textbf{\textcolor{black}{Global alignment.}}
\textcolor{black}{Global molecule--phenotype alignment is performed in the shared space. Paired samples are treated as positives, whereas unpaired samples within the mini-batch serve as negatives~\cite{pattern_recognition:focuscontrast,contrastive:simclr}. After normalization, the similarity between the $i$-th phenotype and the $j$-th molecule is computed as}
\begin{equation}
S_{ij}=
\frac{
(\bar{\boldsymbol{h}}_{v,i}^{s})^{\top}
\bar{\boldsymbol{h}}_{m,j}^{s}
}{\tau},
\end{equation}
\textcolor{black}{where $\tau$ is a learnable temperature parameter. The resulting bidirectional contrastive objective is}
\begin{equation}
\mathcal{L}_{\mathrm{itc}}
=
\frac{1}{2}
\left[
\mathrm{CE}(\mathbf{S},\mathbf{y})
+
\mathrm{CE}(\mathbf{S}^{\top},\mathbf{y})
\right],
\end{equation}
\textcolor{black}{where $\mathbf{y}$ contains the diagonal pairing labels. This objective organizes the shared embedding space so that matched molecule--phenotype pairs are globally closer than unmatched pairs. However, because it compares only global vectors, it does not explicitly assess the compatibility of individual molecule--phenotype pairs after cross-modal interaction.}

\noindent \textbf{\textcolor{black}{Fine-grained matching.}}
\textcolor{black}{To complement global alignment, PhenMol introduces a latent pair-matching module. The shared molecular and cellular representations are treated as single-token sequences and processed by self-attention, bidirectional cross-attention, and feed-forward layers. Because each modality is represented by a single latent token, this module performs instance-level interaction rather than atom-level or image-patch-level co-attention. Let $\boldsymbol{r}_{ij}$ denote the interacted representation of a candidate molecule--phenotype pair. A binary matching head predicts}
\begin{equation}
p_{ij}=\sigma\big(\phi(\boldsymbol{r}_{ij})\big),
\end{equation}
\textcolor{black}{where $p_{ij}$ is the predicted probability that molecule $j$ matches phenotype $i$, $\phi(\cdot)$ is the pair classifier, and $\sigma(\cdot)$ is the sigmoid function. True molecule--phenotype pairs are used as positives, while shuffled pairs provide negatives. The matching loss is}
\begin{equation}
\mathcal{L}_{\mathrm{itm}}
=
-
\frac{1}{B}
\sum_{i=1}^{B}
\left[
y_i \log p_i
+
(1-y_i)\log(1-p_i)
\right],
\end{equation}
\textcolor{black}{where $B$ is the batch size and $y_i \in \{0,1\}$ is the pair label. This module checks pair compatibility after latent interaction and therefore complements the global contrastive objective.}

\subsection{Fine-tuning and downstream prediction}

\textcolor{black}{After pre-training, PhenMol is adapted to downstream tasks using task-specific combinations of the learned shared and private representations.}

\noindent \textbf{\textcolor{black}{Bioactivity prediction.}}
\textcolor{black}{For molecular bioactivity prediction, the model is fine-tuned on a dataset containing more than 16,100 compounds and 270 binary assay tasks. The prediction head concatenates the molecular shared representation, molecular private representation, and paired cellular shared representation:}
\begin{equation}
\hat{\boldsymbol{y}}_i
=
\psi_{\mathrm{bio}}
\left(
[
\boldsymbol{h}^{s}_{m,i};
\boldsymbol{h}^{p}_{m,i};
\boldsymbol{h}^{s}_{v,i}
]
\right),
\end{equation}
\textcolor{black}{where $\psi_{\mathrm{bio}}(\cdot)$ is the bioactivity classifier and $\hat{\boldsymbol{y}}_i$ contains assay-specific activity probabilities. This task evaluates whether phenotype-aware molecular representations improve activity prediction under both random and scaffold splits.}

\noindent \textbf{\textcolor{black}{Modal retrieval.}}
\textcolor{black}{For molecule--phenotype retrieval, no additional classifier is introduced. Molecular and cellular embeddings are extracted from the held-out 10\% paired samples and compared through the full cross-modal similarity matrix:}
\begin{equation}
\mathbf{R}_{ij}
=
\operatorname{sim}
\left(
\boldsymbol{h}^{s}_{v,i},
\boldsymbol{h}^{s}_{m,j}
\right).
\end{equation}
\textcolor{black}{Image-to-molecule retrieval ranks candidate molecules for each phenotype query, whereas molecule-to-image retrieval ranks cellular phenotype profiles for each molecular query. This task directly evaluates cross-modal correspondence learned during pre-training.}

\noindent \textbf{\textcolor{black}{Clinical result prediction.}}
\textcolor{black}{For clinical result prediction, Phase I, Phase II, and Phase III are treated as independent binary classification tasks. Because cellular images are not available at downstream clinical inference, the classifier uses only molecular shared and private representations:}
\begin{equation}
\hat{y}^{(q)}_i
=
\psi^{(q)}_{\mathrm{clin}}
\left(
[
\boldsymbol{h}^{s}_{m,i};
\boldsymbol{h}^{p}_{m,i}
]
\right),
\qquad
q\in\{\mathrm{I},\mathrm{II},\mathrm{III}\}.
\end{equation}
\textcolor{black}{This setting tests whether the learned molecular representation transfers beyond paired Cell Painting readouts to clinically relevant outcome prediction. Together, the three downstream tasks evaluate bioactivity prediction, cross-modal retrieval, and clinical transfer.}

\subsection{Model Training Objective}

PhenMol is trained by jointly optimizing cross-modal alignment, pair-level matching, shared-private decoupling, and modality-specific preservation:
\begin{equation}
\begin{split}
\mathcal{L}_{\mathrm{total}}
=
&\; w_{\mathrm{itc}}\mathcal{L}_{\mathrm{itc}}
+ w_{\mathrm{itm}}\mathcal{L}_{\mathrm{itm}}
+ w_{\mathrm{ortho}}\mathcal{L}_{\mathrm{ortho}} \\
&+ w_{\mathrm{batch}}^{p}\mathcal{L}_{\mathrm{batch}}^{p}
+ w_{\mathrm{batch}}^{s}\mathcal{L}_{\mathrm{batch}}^{s}
+ w_{\mathrm{scaf}}\mathcal{L}_{\mathrm{scaf}} .
\end{split}
\end{equation}
Here, $\mathcal{L}_{\mathrm{itc}}$ and $\mathcal{L}_{\mathrm{itm}}$ promote global and pair-level molecule--phenotype correspondence, $\mathcal{L}_{\mathrm{ortho}}$ reduces redundancy between shared and private representations, the batch-related objectives encourage nuisance variation to remain outside the shared cellular space, and $\mathcal{L}_{\mathrm{scaf}}$ preserves scaffold-level molecular information. All loss weights are fixed during pre-training, while the gradient-reversal coefficient may be progressively increased to stabilize adversarial optimization. The contribution of each objective is examined through ablations on feature decoupling, pair matching, and private-branch supervision.

\section{Experiments}

\subsection{Experimental Setup}

We evaluate \textcolor{black}{PhenMol} along five complementary axes: molecular property prediction, component ablation, molecule--image bidirectional retrieval, clinical trial outcome prediction, and structural preservation. These evaluations test whether the learned representation improves downstream prediction, supports cross-modal matching, transfers to a clinically relevant task, and retains chemically meaningful neighborhoods. 

\noindent \textcolor{black}{\textbf{Pre-training and downstream data.} PhenMol is pre-trained on approximately $3.04 \times 10^{4}$ valid molecule--cell morphology pairs from JUMP-CP and the Broad Drug Repurposing Hub~\cite{dataset:jump_cp,dataset:drug_repurposing_hub}. For evaluation, we use: (i) 270 bioactivity-assay tasks for property prediction~\cite{baseline:moshkov}; (ii) a held-out test library for molecule--image retrieval; and (iii) Phase I--III clinical trial outcomes as independent binary classification tasks following TOP/HINT~\cite{baseline:MIGA,clinical_outcome_assessment:hint}.}

\noindent \textcolor{black}{\textbf{Implementation and training.} SMILES are featurized as 2048-bit ECFP4 fingerprints via RDKit~\cite{rdkit:landrum}, while cellular images are encoded offline via a frozen ResNet-50 into 512-dimensional vectors.}

\noindent \textcolor{black}{\textbf{Evaluation protocols and baselines.} We adopt both random and scaffold splits with a target 8:1:1 train/validation/test ratio. The random split yields 24,323 training, 3,040 validation, and 3,041 test samples. The scaffold split yields 24,985 training, 2,699 validation, and 2,720 test samples; the slight deviation from the target ratio is due to grouping by molecular scaffold. Reporting protocols are task-dependent. For property prediction, we report mean $\pm$ standard deviation over five seeds under random and scaffold splits. Clinical tasks use stratified five-fold cross-validation, and retrieval is evaluated once on the full test library. We compare against a comprehensive set of baselines, including (1) single-modality models trained on Cell Painting profiles~\cite{cellpainting:1,cellpainting:2,baseline:moshkov}, ResNet features~\cite{image:resnet}, MACCS keys~\cite{descriptor:maccs}, and RDKit/Morgan fingerprints~\cite{ecfp:0,rdkit:landrum}; (2) multimodal alignment models such as CLOOME~\cite{baseline:cloome}, MIGA~\cite{baseline:MIGA}, and MINER~\cite{baseline:MINER}; and (3) LR, XGBoost, and HINT for clinical tasks~\cite{baseline:MIGA,clinical_outcome_assessment:hint}. All methods are evaluated under identical splits and protocols, and modality differences in baseline implementations are controlled to ensure fair comparison. For reproducibility, the five property-prediction seeds are 0, 10, 20, 30, and 40; all other task-specific reporting units follow the protocols described above.}

\noindent \textcolor{black}{\textbf{Hardware and main hyperparameters.} Experiments were conducted with PyTorch 2.8.0 and Python 3.12 on Ubuntu 22.04 with CUDA 12.8, using one NVIDIA RTX 4090 GPU with 24GB memory, a 16-vCPU Intel(R) Xeon(R) Gold 6430 CPU, and 120GB RAM. PhenMol uses latent dimension $d=512$ and batch size 64, and is trained for at most 250 epochs. Pre-training uses AdamW with learning rate $4\times10^{-4}$ and weight decay 0.01, while activity-prediction fine-tuning uses AdamW with learning rate $1\times10^{-4}$; retrieval is evaluated without additional optimization. The main loss weights are $w_{\mathrm{itc}}=1.0$, $w_{\mathrm{itm}}=2.0$, $w_{\mathrm{ortho}}=0.5$, $w_{\mathrm{batch}}^p=w_{\mathrm{batch}}^s=0.2$, and $w_{\mathrm{scaf}}=0.2$. The GRL coefficient is fixed to $\alpha=1.0$ by default, and the learnable temperature $\tau$ is initialized to 0.07 and clipped to $[0.001,0.5]$. The downstream classifier takes a $3d=1536$-dimensional input and uses one 256-unit hidden layer with LayerNorm, GELU, and dropout 0.1. Early stopping uses patience 10, monitoring training ITC loss during pre-training and validation mean ROC-AUC during activity-prediction fine-tuning.}

\noindent \textbf{\textcolor{black}{Evaluation metrics.}}
\textcolor{black}{Molecular property prediction is evaluated by AUC, AUP, and \textcolor{black}{AUC$>$0.8/0.85/0.9 thresholds}. Retrieval is evaluated by \textcolor{black}{Hit@1/5/10} and mean reciprocal rank (MRR). Clinical trial outcome prediction is evaluated by AUC and PR-AUC. Structural preservation is evaluated by R-Precision, mean absolute rank shift, and top-$K$ Jaccard using ECFP4-defined molecular neighborhoods.}

\subsection{Molecular Property Prediction}

\begin{table}[t]
\centering
\TableFont
\setlength{\tabcolsep}{3pt}
\caption{Performance comparison on molecular property prediction tasks. Results are reported as mean $\pm$ standard deviation over five seeds. The best results are highlighted in bold, and the second-best results are underlined. The p-value row reports two-sided Welch's t-test results comparing PhenMol with the strongest baseline for each metric within the same split.}
\label{tab:property_prediction}

\resizebox{\columnwidth}{!}{
\begin{tabular}{llccccc}
\toprule
\textbf{Type} & \textbf{Method} & \textbf{AUC} & \textbf{AUP} & 
\textbf{AUC$>$0.8} & \textbf{AUC$>$0.85} & \textbf{AUC$>$0.9} \\
\midrule

\multirow{5}{*}{\shortstack{Single\\mode}}
& CellProfiler & $61.99 \pm 2.3$ & $29.07 \pm 0.5$ & $13.63 \pm 3.4$ & $11.63 \pm 3.5$ & $10.07 \pm 2.9$ \\
& ResNet & $68.43 \pm 1.9$ & $31.70 \pm 2.0$ & $22.96 \pm 0.6$ & $18.67 \pm 0.8$ & $14.89 \pm 0.4$ \\
& MACCS Keys & $63.70 \pm 1.1$ & $26.52 \pm 1.9$ & $18.44 \pm 1.8$ & $13.11 \pm 1.4$ & $9.26 \pm 1.2$ \\
& RDKit FP & $64.69 \pm 1.2$ & $27.78 \pm 2.6$ & $23.93 \pm 1.1$ & $19.70 \pm 1.1$ & $14.96 \pm 1.2$ \\
& Morgan FP & $63.51 \pm 1.7$ & $29.67 \pm 2.1$ & $24.22 \pm 1.2$ & $19.26 \pm 0.0$ & $14.81 \pm 0.5$ \\

\midrule

\multirow{6}{*}{\shortstack{Random\\split}}
& Moshkov & $68.35 \pm 1.9$ & $31.92 \pm 2.7$ & $27.07 \pm 2.2$ & $18.46 \pm 2.1$ & $11.55 \pm 1.4$ \\
& CLOOME & $\underline{69.57 \pm 2.1}$ & $\underline{38.42 \pm 2.1}$ & $26.00 \pm 1.6$ & $\underline{21.63 \pm 2.9}$ & $\underline{15.15 \pm 1.1}$ \\
& MIGA & $63.38 \pm 3.0$ & $27.93 \pm 4.0$ & $\underline{27.57 \pm 3.9}$ & $20.18 \pm 3.2$ & $13.92 \pm 2.5$ \\
& MINER & $66.45 \pm 4.7$ & $31.81 \pm 6.1$ & $20.00 \pm 5.2$ & $14.74 \pm 5.0$ & $10.29 \pm 5.1$ \\
& \textbf{PhenMol} & $\mathbf{73.88 \pm 0.6}$ & $\mathbf{42.09 \pm 0.4}$ & $\mathbf{28.44 \pm 0.7}$ & $\mathbf{22.59 \pm 0.8}$ & $\mathbf{17.63 \pm 1.1}$ \\
& \textit{p-value} & $0.0082$ & $0.0162$ & $0.0231$ & $0.5100$ & $0.0074$ \\

\midrule

\multirow{6}{*}{\shortstack{Scaffold\\split}}
& Moshkov & $\underline{66.15 \pm 2.3}$ & $\underline{31.06 \pm 2.9}$ & $22.81 \pm 2.0$ & $\underline{17.26 \pm 1.9}$ & $\underline{10.89 \pm 1.7}$ \\
& CLOOME & $65.75 \pm 0.2$ & $28.72 \pm 0.2$ & $\underline{25.11 \pm 0.7}$ & $13.85 \pm 0.6$ & $6.67 \pm 0.3$ \\
& MIGA & $55.68 \pm 1.1$ & $20.62 \pm 0.6$ & $10.28 \pm 1.0$ & $7.71 \pm 1.1$ & $4.90 \pm 0.7$ \\
& MINER & $64.33 \pm 1.4$ & $25.09 \pm 1.2$ & $18.74 \pm 1.3$ & $12.00 \pm 1.6$ & $8.81 \pm 0.9$ \\
& \textbf{PhenMol} & $\mathbf{72.01 \pm 1.0}$ & $\mathbf{40.00 \pm 0.6}$ & $\mathbf{26.81 \pm 1.1}$ & $\mathbf{22.00 \pm 0.8}$ & $\mathbf{17.70 \pm 0.8}$ \\
& \textit{p-value} & $0.0324$ & $0.0291$ & $0.0536$ & $0.0346$ & $0.0095$ \\

\bottomrule
\end{tabular}
}
\end{table}

\textcolor{black}{We conducted experiments on 270 bioactivity-assay tasks to test the model generalization under both random and scaffold splits. As shown in Table~\ref{tab:property_prediction}, PhenMol achieves the best performance across all reported metrics under both random and scaffold splits. Under the random split, it improves AUC from $69.57$ to \textbf{$73.88$ (+4.31)} and AUP from $38.42$ to \textbf{$42.09$ (+3.67)} compared with the strongest multimodal baseline (i.e., CLOOME). The gains are statistically significant with $p<0.05$ for AUC and AUP, respectively. Under the more challenging scaffold split, PhenMol maintains an AUC of $72.01$ and AUP of $40.00$, outperforming the strongest baseline by $5.86\%$ and $8.94\%$, respectively, and this advantage grows substantially under scaffold split. This widening gap suggests that structure-preserving alignment confers particular benefits when the model must generalize beyond its training chemical distribution.}

\textcolor{black}{The differential improvements in AUC and AUP can be attributed to our model's enhanced integration of multimodal data, while scaffold-split results reveal the importance of preserving chemical structure. The AUC gain reflects improved overall ranking across active and inactive compounds, consistent with the use of shared phenotype information and latent pair matching to inject biological response signals. The larger AUP gain is diagnostically significant for drug discovery, as AUP emphasizes performance on rare positive examples, and this capability is critical for hit identification. The scaffold split offers a more rigorous test because test molecules are chemically distinct from training molecules at the scaffold level. PhenMol's larger margin in this setting underscores a key design principle. For phenotype-aware molecular representation learning, the critical challenge is not simply whether modalities are aligned, but whether the alignment preserves chemically transferable structure under scaffold shift. The scaffold-preserving private branch appears to serve this function by maintaining chemical identity that might otherwise be overwritten by phenotype-specific noise during alignment. At the same time, the observation that several phenotype-aware baselines do not consistently exceed single-modality performance suggests that global cross-modal alignment can import noise or weaken molecular structure when two modalities are forced into a single geometry without appropriate structural constraints.}

\subsection{Ablation Study}

\begin{figure}[t]
    \centering
    \includegraphics[width=\columnwidth]{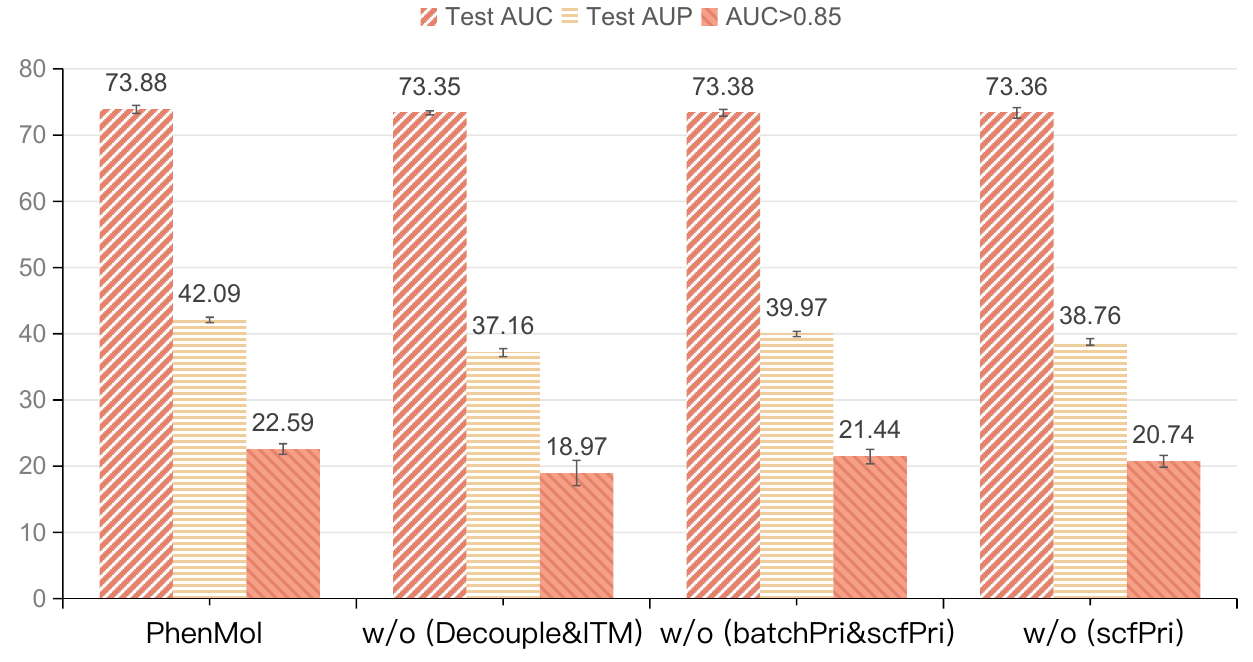}
    \caption{Ablation study of key modules in PhenMol.}
    \label{fig:ablation}
\end{figure}

\textcolor{black}{We conducted ablation studies on the 270 bioactivity-assay tasks to investigate factors influencing performance.} \textcolor{black}{\figref{fig:ablation} reports results for different model variants: \textbf{PhenMol w/o Decouple\&ITM} removes both feature decoupling and image--molecule matching; \textbf{PhenMol w/o batchPri\&scfPri} removes the private-branch auxiliary supervision for image batch prediction and molecular scaffold prediction; \textbf{PhenMol w/o scfPri} removes only the molecular scaffold prediction constraint; and PhenMol denotes the complete model.}

\textcolor{black}{The ablation results show that PhenMol's gain comes from the interaction among decoupling, pair-level matching, and private-branch constraints rather than from a single auxiliary loss. Removing feature decoupling slightly increases average AUC, from $73.88$ to $74.99$, but reduces AUP from $42.09$ to $39.31$ and lowers high-AUC task coverage. This discrepancy indicates that average AUC alone can overstate performance when positive compounds are sparse: the model may rank negatives well overall while becoming less reliable at identifying rare active compounds. Removing ITM produces a similar but milder pattern, showing that global contrastive alignment alone is not sufficient for stable pair-level discrimination.}

\textcolor{black}{The strongest degradation appears when decoupling and ITM are removed together, reducing AUP to $37.16$ and lowering the proportion of tasks with AUC $>0.8$ to $24.79\%$. This result supports the design in \figref{fig:modal_thir}: global alignment and fine-grained matching are complementary, but they need a decoupled representation space so that phenotype information does not overwrite molecular structure. Removing private-branch constraints also reduces AUP, especially when scaffold prediction is removed. This confirms the role of the molecular private branch as a structure-preserving path rather than a redundant auxiliary head.}

\subsection{Molecule--Image Retrieval}

\begin{table*}[t]
\centering
\TableFont
\setlength{\tabcolsep}{4pt}
\caption{Molecule--image bidirectional retrieval results under random and scaffold splits. Hit@K is reported in percentage. MRR denotes mean reciprocal rank. Retrieval is evaluated over the complete test candidate library rather than mini-batches. The best results are highlighted in bold, and the second-best results are underlined.}
\label{tab:retrieval}

\begin{tabular*}{\textwidth}{@{\extracolsep{\fill}}lllcccc@{}}
\toprule
\textbf{Split} & \textbf{Direction} & \textbf{Method} & 
\textbf{Hit@1} & \textbf{Hit@5} & \textbf{Hit@10} & \textbf{MRR} \\
\midrule

\multirow{12}{*}{Random}
& Image-to-molecule & Random & 0.03 & 0.16 & 0.33 & 0.0000 \\
& Image-to-molecule & InfoNCE & 10.03 & 29.10 & 38.84 & 0.1967 \\
& Image-to-molecule & InfoLOOB & \underline{10.70} & \underline{30.94} & \underline{41.85} & \underline{0.2054} \\
& Image-to-molecule & MIGA & 8.85 & 25.72 & 36.80 & 0.1756 \\
& Image-to-molecule & MINER & 8.11 & 24.05 & 35.13 & 0.1447 \\
& Image-to-molecule & \textbf{PhenMol} & \textbf{11.74} & \textbf{33.08} & \textbf{45.38} & \textbf{0.2227} \\

\cmidrule(lr){2-7}

& Molecule-to-image & Random & 0.03 & 0.16 & 0.33 & 0.0000 \\
& Molecule-to-image & InfoNCE & 10.75 & 29.43 & 38.01 & 0.2011 \\
& Molecule-to-image & InfoLOOB & \underline{11.06} & \underline{30.22} & \underline{42.76} & \underline{0.2059} \\
& Molecule-to-image & MIGA & 8.68 & 26.37 & 36.67 & 0.1755 \\
& Molecule-to-image & MINER & 8.21 & 24.59 & 35.36 & 0.1645 \\
& Molecule-to-image & \textbf{PhenMol} & \textbf{12.56} & \textbf{33.74} & \textbf{44.33} & \textbf{0.2259} \\

\midrule

\multirow{12}{*}{Scaffold}
& Image-to-molecule & Random & 0.04 & 0.18 & 0.37 & 0.0000 \\
& Image-to-molecule & InfoNCE & 0.95 & 4.26 & \underline{9.59} & 0.0400 \\
& Image-to-molecule & InfoLOOB & 0.77 & 4.78 & 9.41 & 0.0406 \\
& Image-to-molecule & MIGA & 0.97 & 4.68 & 9.11 & 0.0426 \\
& Image-to-molecule & MINER & \underline{1.06} & \underline{4.98} & 9.51 & \underline{0.0446} \\
& Image-to-molecule & \textbf{PhenMol} & \textbf{1.14} & \textbf{6.10} & \textbf{11.40} & \textbf{0.0482} \\

\cmidrule(lr){2-7}

& Molecule-to-image & Random & 0.03 & 0.16 & 0.33 & 0.0000 \\
& Molecule-to-image & InfoNCE & 0.94 & 4.04 & 8.33 & 0.0426 \\
& Molecule-to-image & InfoLOOB & 1.10 & 4.74 & 9.04 & 0.0424 \\
& Molecule-to-image & MIGA & 0.99 & 4.50 & 8.70 & 0.0421 \\
& Molecule-to-image & MINER & \underline{1.18} & \underline{5.04} & \underline{9.41} & \underline{0.0458} \\
& Molecule-to-image & \textbf{PhenMol} & \textbf{1.36} & \textbf{5.33} & \textbf{9.82} & \textbf{0.0481} \\

\bottomrule
\end{tabular*}

\end{table*}

\textcolor{black}{We further evaluate cross-modal alignment through bidirectional retrieval. Image-to-molecule retrieval asks whether a cellular phenotype can recover its inducing molecule, whereas molecule-to-image retrieval asks whether a molecule can recover its paired cellular response. This task directly tests whether the learned representation places paired molecules and phenotypes close to one another, with results summarized in Table~\ref{tab:retrieval}.}

\begin{figure*}[t]
    \centering
    \includegraphics[width=\textwidth]{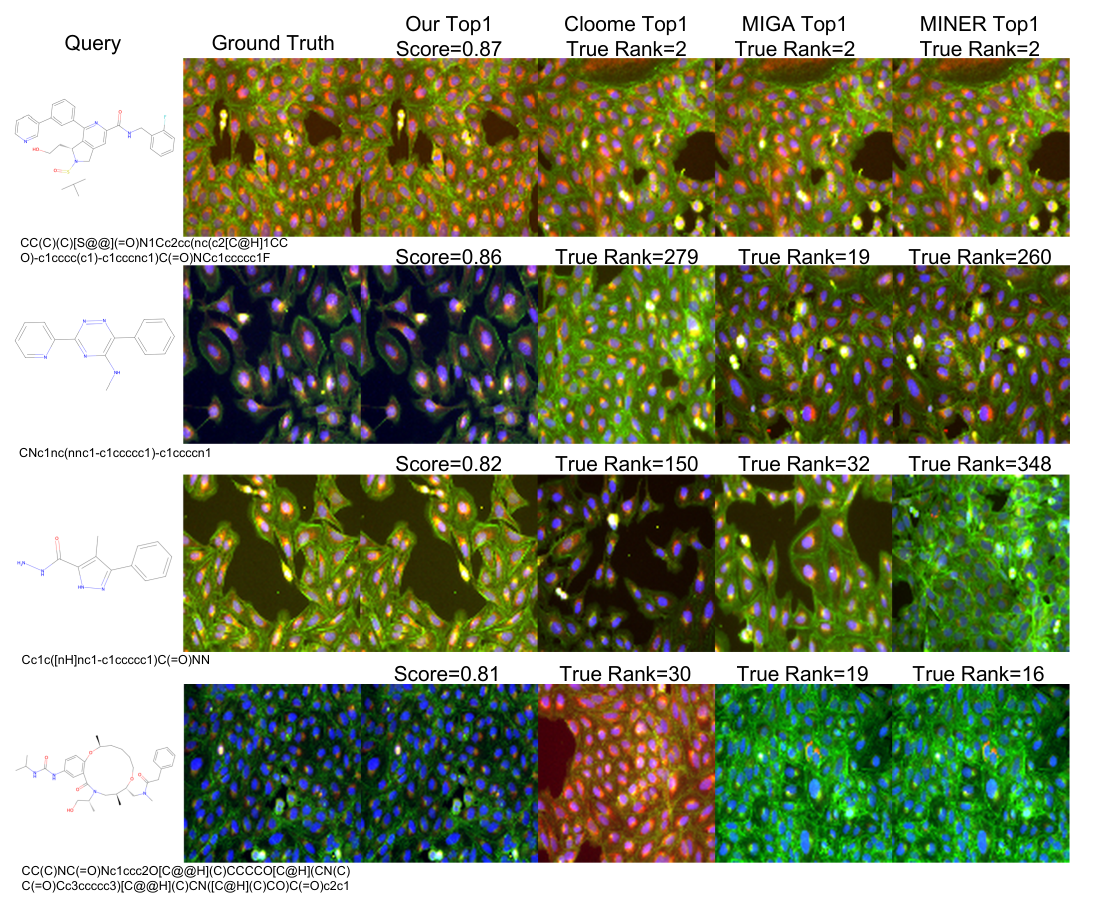}
    \caption{Qualitative scaffold-split example of molecule--image retrieval under competing multimodal methods. Given a query sample and its ground-truth match, the figure shows the Top-1 retrieval result produced by \textcolor{black}{PhenMol} , CLOOME, MIGA, and MINER. The retrieval score for \textcolor{black}{PhenMol}  and the ground-truth rank assigned by each baseline are reported to illustrate whether each method retrieves the correct paired sample near the top of the full test-set ranking.}
    \label{fig:retrieval_example}
\end{figure*}

To complement the aggregate retrieval metrics, we visualize a representative scaffold-split retrieval case in \figref{fig:retrieval_example}. The example compares the Top-1 retrieval result of each method and reports where the ground-truth pair appears in each method's ranked list.

Under the random split, \textcolor{black}{PhenMol}  achieves the best performance in both retrieval directions. For image-to-molecule retrieval, it obtains 11.74\% Hit@1, 33.08\% Hit@5, 45.38\% Hit@10, and 0.2227 MRR, consistently outperforming InfoNCE, InfoLOOB, MIGA, and MINER. For molecule-to-image retrieval, \textcolor{black}{PhenMol}  also achieves the highest Hit@1, Hit@5, Hit@10, and MRR scores. \textcolor{black}{The comparison with InfoNCE and InfoLOOB is informative: these conventional contrastive objectives are competitive under the random split, indicating that global alignment can work when test molecules are close to training molecules. However, they only optimize vector-level similarity and do not explicitly separate phenotype-relevant information from chemical identity or nuisance variation.}

Under the scaffold split, retrieval becomes substantially more challenging for all methods, reflecting the difficulty of cross-modal matching on unseen chemical scaffolds. \textcolor{black}{The absolute Hit@1 values drop sharply, showing that cellular phenotypes alone do not make scaffold-level retrieval easy. This is the setting where naive multimodal alignment is most vulnerable: images may capture similar morphology for chemically different perturbations, while scaffold changes can move the true molecular match far from the training distribution. PhenMol improves this setting by coupling global contrastive alignment with fine-grained latent pair matching and structure-preserving decoupling. The contrastive objective organizes the coarse image--molecule space, the matching branch checks pair compatibility after latent interaction, and the private molecular branch prevents the shared alignment objective from erasing scaffold-level information.This experiment emphasizes the importance of combining multimodal alignment with chemical-structure preservation rather than treating retrieval as a pure contrastive matching problem.}

\begin{figure*}[t]
    \centering
    \includegraphics[width=\textwidth]{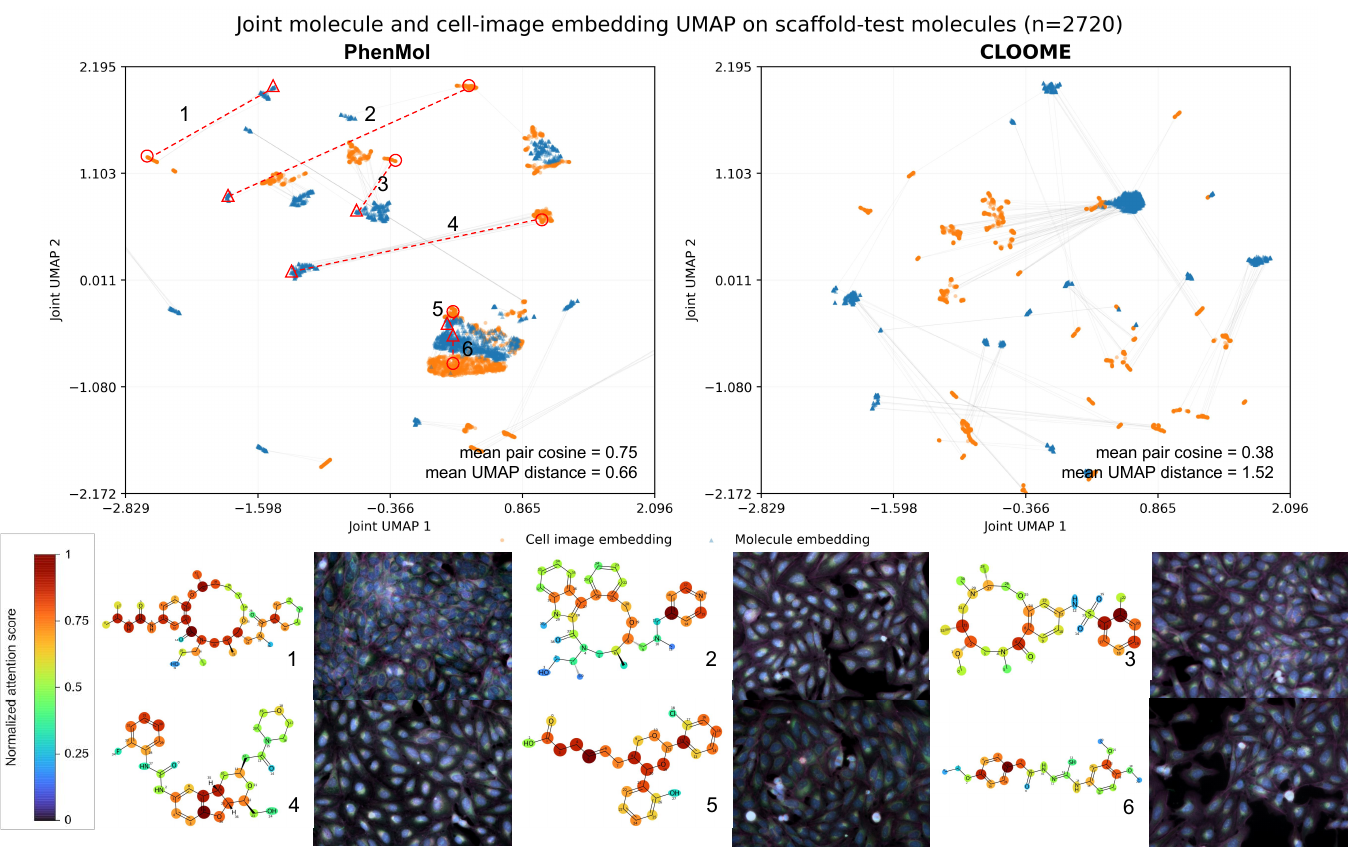}
    \caption{Visualization of image--molecule alignment and molecular attention on scaffold-test retrieval examples. Orange points denote cellular phenotype embeddings, blue triangles denote molecular embeddings, and gray lines connect paired observations. The lower panels show selected molecule--image pairs, with atom color indicating normalized attention score.}
    \label{fig:attention_alignment}
\end{figure*}

\textcolor{black}{We further inspect the paired molecule--cell embedding space on the scaffold-test molecules used for retrieval evaluation in \figref{fig:attention_alignment}. PhenMol places paired cell-image and molecule embeddings closer than CLOOME in the displayed joint projection, with a mean pair cosine similarity of 0.75 and a mean UMAP distance of 0.66, compared with 0.38 and 1.52 for CLOOME. These values are descriptive summaries of the visualization rather than separate evaluation metrics. The lower panels show selected retrieved molecule--image pairs with normalized atom-level attention scores, which should be interpreted as model-derived saliency over the molecular graph rather than direct evidence of a biochemical mechanism.}

\subsection{Clinical Trial Outcome Prediction}

\textcolor{black}{We evaluate whether the learned molecular representations transfer beyond activity assays to clinical trial outcome prediction. This task tests whether phenotype-aware molecular representations contain information relevant to real-world drug development outcomes, with results summarized in Table~\ref{tab:clinical_prediction}.}

\begin{table*}[t]
\centering
\TableFont
\setlength{\tabcolsep}{5pt}
\caption{Performance comparison on clinical trial outcome prediction. Results are reported as mean $\pm$ standard deviation across stratified five-fold cross-validation within each phase. The best results are highlighted in bold, and the second-best results are underlined.}
\label{tab:clinical_prediction}

\begin{tabular*}{0.9\columnwidth}{@{\extracolsep{\fill}}llcc@{}}
\toprule
\textbf{Method} & \textbf{Phase} & \textbf{AUC} & \textbf{PR-AUC} \\
\midrule

\multirow{3}{*}{LR}
& Phase I & $0.487 \pm 0.006$ & $0.634 \pm 0.007$ \\
& Phase II & $0.534 \pm 0.017$ & $0.509 \pm 0.014$ \\
& Phase III & $0.528 \pm 0.003$ & $0.675 \pm 0.010$ \\

\midrule

\multirow{3}{*}{XGBoost}
& Phase I & $0.508 \pm 0.006$ & $0.646 \pm 0.003$ \\
& Phase II & $0.516 \pm 0.007$ & $0.481 \pm 0.004$ \\
& Phase III & $0.597 \pm 0.015$ & $0.712 \pm 0.009$ \\

\midrule

\multirow{3}{*}{HINT}
& Phase I & $0.516 \pm 0.005$ & $\underline{0.673 \pm 0.015}$ \\
& Phase II & $0.584 \pm 0.003$ & $0.537 \pm 0.004$ \\
& Phase III & $0.621 \pm 0.006$ & $0.689 \pm 0.003$ \\

\midrule

\multirow{3}{*}{CLOOME}
& Phase I & $0.514 \pm 0.024$ & $0.605 \pm 0.005$ \\
& Phase II & $0.563 \pm 0.014$ & $0.491 \pm 0.021$ \\
& Phase III & $0.571 \pm 0.033$ & $0.724 \pm 0.036$ \\

\midrule

\multirow{3}{*}{MIGA}
& Phase I & $\underline{0.569 \pm 0.050}$ & $0.659 \pm 0.040$ \\
& Phase II & $\mathbf{0.593 \pm 0.012}$ & $\mathbf{0.563 \pm 0.017}$ \\
& Phase III & $0.603 \pm 0.039$ & $\underline{0.757 \pm 0.024}$ \\

\midrule

\multirow{3}{*}{MINER}
& Phase I & $0.564 \pm 0.033$ & $0.633 \pm 0.037$ \\
& Phase II & $0.581 \pm 0.012$ & $0.515 \pm 0.016$ \\
& Phase III & $\underline{0.631 \pm 0.043}$ & $0.751 \pm 0.038$ \\

\midrule

\multirow{3}{*}{\textbf{PhenMol}}
& Phase I & $\mathbf{0.616 \pm 0.010}$ & $\mathbf{0.675 \pm 0.040}$ \\
& Phase II & $\mathbf{0.593 \pm 0.020}$ & $\underline{0.521 \pm 0.026}$ \\
& Phase III & $\mathbf{0.669 \pm 0.037}$ & $\mathbf{0.778 \pm 0.029}$ \\

\bottomrule
\end{tabular*}

\end{table*}

\textcolor{black}{PhenMol achieves the best AUC and PR-AUC in Phase I, with an AUC of $0.616$ and a PR-AUC of $0.675$. It also achieves the best Phase III performance, with an AUC of $0.669$ and a PR-AUC of $0.778$. These two phases represent different clinical settings: early-stage trials emphasize safety and tolerability signals, whereas Phase III trials are closer to efficacy confirmation in larger populations. The improvement across both phases indicates that the learned molecular representation contains transferable information beyond the original cell-imaging pre-training task.}

\textcolor{black}{In Phase II, PhenMol matches the best AUC but has lower PR-AUC than MIGA. This mixed Phase II result provides an important boundary. Mid-stage clinical outcomes are strongly influenced by indication, patient selection, dose, endpoint design, and protocol context, which are not fully captured by molecule--phenotype representation learning. Thus, PhenMol improves the molecular component of clinical transfer, but the result also shows that molecule and cellular phenotype information alone cannot replace richer trial-level modeling.}

\subsection{Structural Preservation Analysis}
\begin{figure*}[t]
    \centering
    \includegraphics[width=\textwidth]{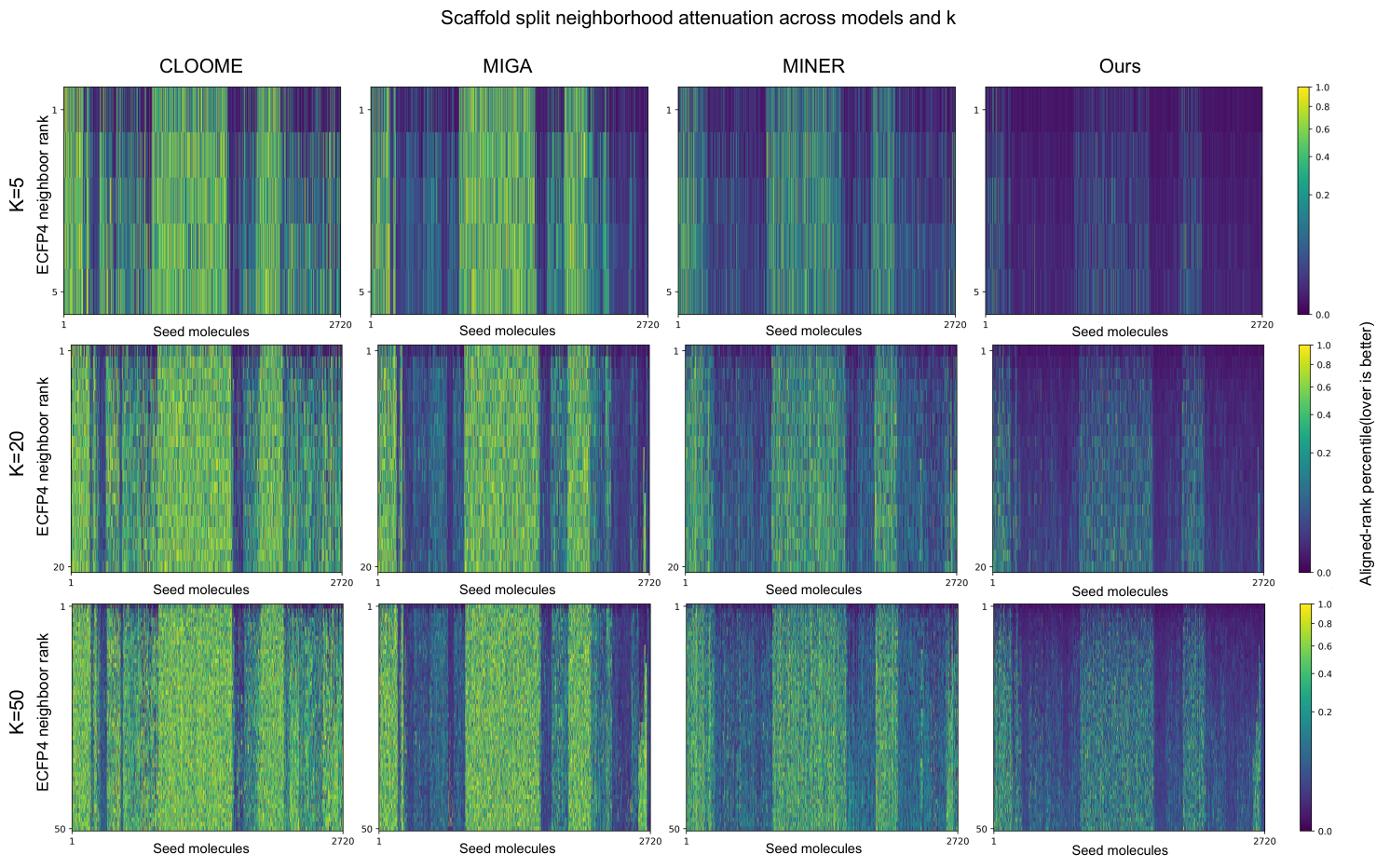}
    \caption{Visualization of molecular structural neighborhood preservation in different multimodal embedding spaces. Compared with global alignment methods such as CLOOME and robustness-enhanced methods such as MIGA and MINER, \textcolor{black}{PhenMol}  better maintains ECFP4-defined structural neighbors after multimodal learning.}
    \label{fig:structure_visualization}
\end{figure*}

\textcolor{black}{Finally, we examine whether PhenMol preserves molecular structural neighborhoods after multimodal learning. This analysis directly tests the central concern that cross-modal alignment can distort chemically meaningful molecular organization. \figref{fig:structure_visualization} provides a qualitative comparison of learned embedding spaces, and Table~\ref{tab:structure_preservation} quantifies whether ECFP4-defined neighbors remain close after multimodal training.}

\begin{table}[t]
\centering
\TableFont
\setlength{\tabcolsep}{5pt}
\caption{Structural preservation analysis based on ECFP4-defined molecular neighborhoods.}
\label{tab:structure_preservation}

\begin{tabular*}{0.75\columnwidth}{@{\extracolsep{\fill}}ccccc@{}}
\toprule
\textbf{$K$} & \textbf{Model} & 
\textbf{R-Prec.$\uparrow$} & 
\textbf{Rank Shift$\downarrow$} & 
\textbf{Jaccard$\uparrow$} \\
\midrule

\multirow{4}{*}{5}
& MINER   & 0.1766 & 298.2 & 0.1137 \\
& MIGA    & 0.1435 & 562.4 & 0.0958 \\
& CLOOME  & 0.1427 & 708.9 & 0.0950 \\
& \textbf{PhenMol} & \textbf{0.4480} & \textbf{61.8} & \textbf{0.3313} \\

\midrule

\multirow{4}{*}{20}
& MINER   & 0.1914 & 390.5 & 0.1166 \\
& MIGA    & 0.1442 & 606.9 & 0.0886 \\
& CLOOME  & 0.1150 & 853.2 & 0.0669 \\
& \textbf{PhenMol} & \textbf{0.4552} & \textbf{131.3} & \textbf{0.3230} \\

\midrule

\multirow{4}{*}{50}
& MINER   & 0.2205 & 459.2 & 0.1432 \\
& MIGA    & 0.2211 & 649.5 & 0.1331 \\
& CLOOME  & 0.1307 & 917.8 & 0.0758 \\
& \textbf{PhenMol} & \textbf{0.4563} & \textbf{209.8} & \textbf{0.3195} \\

\bottomrule
\end{tabular*}

\end{table}

\textcolor{black}{PhenMol}  consistently achieves the best R-Precision and top-$K$ Jaccard, together with the lowest mean absolute rank shift across all neighborhood sizes. At $K=5$, \textcolor{black}{PhenMol}  improves R-Precision from $0.1766$ to $0.4480$ compared with the strongest baseline and reduces mean rank shift from $298.2$ to $61.8$. \textcolor{black}{This means that close ECFP4 neighbors remain much closer in the learned representation, instead of being displaced by phenotype-driven alignment.}
The advantage remains at larger neighborhood sizes, with R-Precision of $0.4552$ at $K=20$ and $0.4563$ at $K=50$. \textcolor{black}{This scale consistency is important because local neighborhood preservation and broader scaffold-level organization are different requirements: a method can preserve a few nearest analogs while still distorting the larger chemical neighborhood. PhenMol improves both, indicating that the private scaffold-preserving branch constrains molecular geometry beyond the immediate nearest neighbors. Together with the property prediction and retrieval results, this analysis supports the central mechanism of the paper: phenotype-aware learning is most effective when cross-modal alignment is coupled with explicit molecular structure preservation.}

\begin{figure*}[t]
    \centering
    \includegraphics[width=\textwidth]{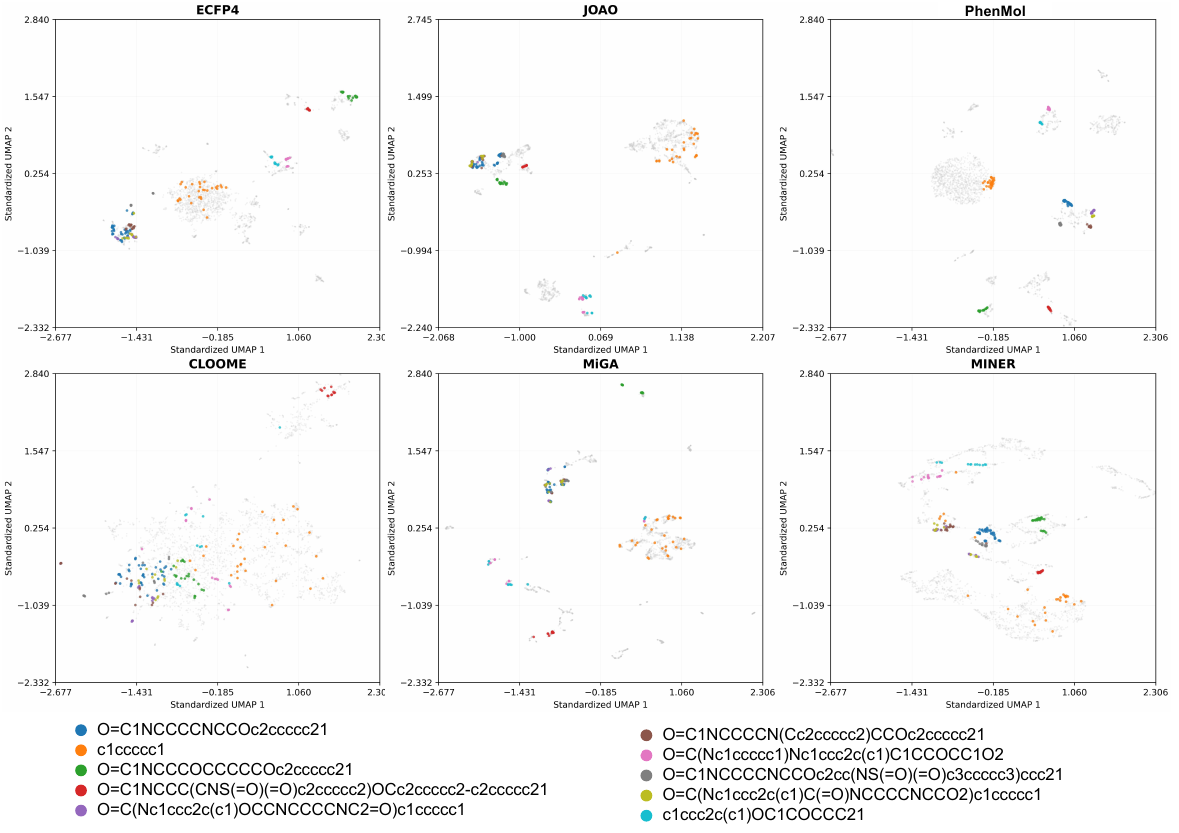}
    \caption{UMAP visualization of molecular embedding spaces on the scaffold-test molecules. Gray points denote the full molecule set, and colored points highlight representative molecules from related chemical neighborhoods.}
    \label{fig:umap_structure}
\end{figure*}

\textcolor{black}{The UMAP visualization in \figref{fig:umap_structure} provides a complementary view of the scaffold-test molecular embedding space. Gray points denote the full molecule set, and colored points highlight representative molecules from related chemical neighborhoods. Compared with multimodal alignment baselines whose local organization becomes more diffuse, PhenMol maintains compact highlighted neighborhoods while still being trained with paired cellular phenotype information, indicating that structure preservation is visible in the learned geometry rather than only in the rank-based metrics.}

\section{Conclusion}

\textcolor{black}{We introduced PhenMol, a novel approach designed to address a key failure mode in phenotype-aware molecular representation learning: cellular image supervision can improve biological alignment but may also distort chemically meaningful molecular neighborhoods. PhenMol incorporates structure-preserving multimodal learning that constrains cross-modal alignment by chemical structure rather than optimizing solely for global image--molecule similarity. Extensive experiments show that PhenMol consistently outperforms state-of-the-art methods under both random and scaffold splits, achieving superior accuracy and robustness in molecular property prediction, bidirectional molecule--phenotype retrieval, and clinical trial outcome prediction. PhenMol also preserves ECFP4-defined structural neighborhoods more effectively than existing cross-modal methods and demonstrates strong Phase I and Phase III clinical trial results, offering new insights into the role of chemical structure preservation in phenotype-aware molecular learning and facilitating the discovery of novel therapeutic solutions.}

\textcolor{black}{PhenMol} learns from molecular fingerprints and pre-extracted cellular phenotype features, but does not explicitly model targets, pathways, transcriptomic responses, disease context, or clinical trial variables. As a result, although it improves structure-preserving molecule--cell alignment, its predictions cannot fully explain the underlying biological mechanisms or capture all factors that determine clinical outcomes, especially in Phase II trials where indication, dosage, patient population, endpoints, and protocol design are often critical. Therefore, \textcolor{black}{PhenMol} should be used as a research-oriented representation-learning and candidate-prioritization tool rather than a standalone system for clinical decision-making. Its predictions should be treated as hypotheses requiring experimental validation and, where relevant, independent clinical assessment. Future work should incorporate richer biomedical and clinical modalities and improve interpretability by linking molecular substructures to cellular phenotypic patterns and downstream outcomes.

\section*{\textcolor{black}{CRediT Authorship Contribution Statement}}

\textcolor{black}{Xuan Lin: Resources, Methodology, Investigation, Formal analysis, Funding acquisition, Conceptualization. Jingyu Sheng: Writing -- original draft, Visualization, Validation, Methodology, Investigation, Data curation, Conceptualization. Tengfei Ma: Visualization, Supervision, Project administration, Investigation, Conceptualization, Funding acquisition. Li Sun: Writing -- review \& editing, Resources, Methodology, Investigation, Formal analysis. Dapeng Xiong: Project administration, Investigation, Data curation, Funding acquisition.}

\section*{Acknowledgements}
We thank our colleagues for their helpful discussions and constructive suggestions. This research was supported by the National Natural Science Foundation of China (No. 62573372 to Xuan Lin and No. 625B2067 to Tengfei Ma), and by the Southeast University Startup Grant (No. 4007032510 to D.X.).
\section*{Declaration of Generative AI and AI-Assisted Technologies in the Writing Process}
Generative AI tools were used to assist with language editing, manuscript organization, and clarity checks during the preparation of this manuscript. They were not used to generate experimental data, run analyses, create evaluation results, or make autonomous scientific conclusions.

\section*{\textcolor{black}{Data Availability}}

\textcolor{black}{The code and data used in this study are available at \url{https://github.com/JacklinGroup/Phenmol}.}

\bibliographystyle{elsarticle-num-nourl}
\bibliography{references}

\end{document}